\documentclass[conference]{IEEEtran}
\pdfoutput=1
\IEEEoverridecommandlockouts

\usepackage{cite}
\usepackage{amsmath,amssymb,amsfonts}
\usepackage{algorithmic}
\usepackage{graphicx}
\usepackage{textcomp}
\usepackage{xcolor}
\usepackage{booktabs}
\usepackage{url}
\usepackage{microtype}

\def\BibTeX{{\rm B\kern-.05em{\sc i\kern-.025em b}\kern-.08em
    T\kern-.1667em\lower.7ex\hbox{E}\kern-.125emX}}

\begin{document}

\title{Lifecycle-Optimal Tokenization: Vocabulary Size as a
Deployment-Regime-Dependent Infrastructure Parameter}

\author{
\IEEEauthorblockN{Rima Mittal}
\IEEEauthorblockA{rimamittal@gmail.com}
\and
\IEEEauthorblockN{Ankit Gubrani}
\IEEEauthorblockA{ankit.gubrani90@gmail.com}
\and
\IEEEauthorblockN{Satyanarayana Kakollu}
\IEEEauthorblockA{kakollu@icloud.com}
}

\maketitle

\begin{abstract}
Tokenizer vocabulary size is a foundational design choice in large language model (LLM)
infrastructure, yet it is typically fixed at training time based on convention rather than
deployment analysis. We show that the cost-optimal vocabulary is not a constant but a
function of the serving regime. We formalize total deployment cost as
$C_{\text{lifecycle}}(V) = C_{\text{train}}(V) + \lambda \cdot C_{\text{infer}}(V, B)$,
where $\lambda$ is inference volume and $B$ is the serving batch size.
Through controlled experiments on two GPU families spanning the memory-bound to
compute-bound regimes (A10G, ridge $\approx$117 FLOP/byte; A100, ridge $\approx$183
FLOP/byte), we demonstrate: (1) the inference-optimal vocabulary shifts 16$\times$ with
serving batch, from 32k at $B=1$ to 524k at $B=64+$, driven by amortization of
the $V \times d$ unembedding matrix read; (2) at 1.3--2.3B model scale, quality (bits
per byte, BPB) is optimized at $V=65$k, confirming scale-dependent vocabulary preference;
(3) the lifecycle-optimal vocabulary diverges from training-optimal by up to 16$\times$
for production deployments. Quality is approximately invariant across the optimal range
($<$2\% BPB spread), making vocabulary a pure systems optimization with no quality
penalty in the measured range. Our results provide actionable capacity planning
guidance: on-device deployments ($B=1$) should use $V \approx 32$k; datacenter serving
($B \geq 64$, $\lambda \geq 10$) should use $V \approx 131$--262k.
\end{abstract}


\section{Introduction}

Every large language model deployed in production carries a vocabulary -- a fixed set of
subword tokens used to encode all input and output text. Vocabulary size $V$ is set once
at tokenizer training time, before any inference infrastructure is planned.
Yet $V$ has first-order effects on both training throughput and inference cost: it
determines the size of the unembedding matrix $W \in \mathbb{R}^{V \times d}$ that
must be read from high-bandwidth memory (HBM) at every decode step.

At each autoregressive step, the model produces a $d$-dimensional hidden vector and
projects it through $W$ to obtain logits over all $V$ tokens. This projection dominates
decode-time memory traffic: for vocabulary $V=262144$ and hidden dimension $d=512$ in
fp16, the weight matrix alone is $262144 \times 512 \times 2 = 256$\,MB read per step.
The cost of this read depends critically on the serving batch size $B$.

The key physical insight follows from roofline analysis~\cite{roofline}.
The unembedding matmul $[\mathbf{h}]_{B \times d} \cdot [W^T]_{d \times V}$ has
arithmetic intensity $I \approx B$ FLOP/byte (fp16). At $B=1$ (on-device),
$I = 1 \ll \rho$ where $\rho$ is the GPU's ridge point, so the operation is
memory-bound: cost scales linearly with $V$ and the weight matrix is streamed at
near-peak bandwidth. At $B=256$ (datacenter), $I = 256 > \rho$, the operation is
compute-bound: the same $V \times d$ read is amortized across 256 sequences,
reducing per-sequence cost by $\sim$256$\times$.
The \emph{optimal vocabulary therefore depends on the serving batch size} --
a deployment parameter determined long after the tokenizer is frozen.

This observation has direct capacity planning implications. Infrastructure teams
provisioning GPU clusters for LLM serving must account for how vocabulary choice
affects memory bandwidth requirements, throughput per accelerator, and the number
of accelerators needed to meet latency SLAs. A model with a suboptimally small
vocabulary wastes GPU cycles at datacenter batch; a model with a suboptimally large
vocabulary degrades on-device latency. Neither is captured by existing vocabulary
selection methods, which optimize for training quality~\cite{tao2024} or single-sample
inference~\cite{lengthmax} in isolation.

Current practice selects $V$ from convention (32k in Llama-2/3~\cite{llama2},
100k+ in GPT-4~\cite{gpt4}, 256k in Gemma-2~\cite{gemma2}) or from training-time
quality metrics~\cite{tao2024}. Neither accounts for how the cost of the unembedding
step changes with the serving regime.

\subsection{Contributions}

\begin{enumerate}
\item A \textbf{lifecycle cost model} $C_\text{lifecycle}(V, B, \lambda)$ that jointly
captures training and inference costs over vocabulary size, parameterized by serving
batch $B$ and inference volume $\lambda$. Both cost components share physical units
(ms/byte), making $\lambda$ a dimensionless, physically interpretable ratio.

\item \textbf{Empirical measurement} of the inference-optimal vocabulary across batch
sizes $B \in \{1, 16, 64, 256, 1024, 4096\}$ on two GPU ridge points (A10G
$\approx$117, A100 $\approx$183 FLOP/byte), with kernel-launch overhead eliminated
via CUDA graph capture and isolated head measurement. The 16$\times$ shift in optimal
$V$ is confirmed overhead-free.

\item \textbf{Scale-dependent quality measurement} at 1.3--2.3B parameters using
Fully Sharded Data Parallel (FSDP) with bf16 precision, showing the quality-optimal
vocabulary shifts from 16k (at 100M) to 65k (at 1.5B), consistent with Tao et al.'s
$V^* \propto N^{0.5}$ scaling law~\cite{tao2024}.

\item A \textbf{lifecycle sweep table} (Table~\ref{tab:lifecycle}) and
\textbf{provisioning recommendations} providing direct guidance as a function of
$\lambda$ and $B$, suitable for infrastructure capacity planning decisions.
\end{enumerate}

\section{Background and Related Work}

\subsection{Prior Work on Vocabulary Optimization}

\textbf{Tao et al. (NeurIPS 2024)}~\cite{tao2024} derive that the loss-optimal
vocabulary scales as $V^* \propto N^{0.5}$ where $N$ is model parameters. Their
analysis is FLOPs-only at training time; no inference cost, batch dependence,
or hardware regime is modeled. We extend their quality insight with an inference
cost model parameterized by serving batch.

\textbf{Length-MAX (2025)}~\cite{lengthmax} builds inference-efficient tokenizers
by optimizing $\text{score}(t) = \text{freq}(t) \cdot |t|$ to balance token frequency
and length. Benchmarks are on A100 at $B=1$ only; no batch sweep, no lifecycle
framework, and the cost function contains no bytes-moved term.

\textbf{Compute-Optimal Tokenization (2026)}~\cite{cot2026} addresses training-time
FLOPs combined with compression and loss (BPB). Their Section 3.5 explicitly states
``the inference-optimal frontier [is] underexplored.'' We address this gap directly.

\textbf{Hardware Co-Design Scaling Laws (2026)}~\cite{roofline2026} co-designs
depth, width, and quantization against a roofline model on Jetson Orin, but holds
vocabulary constant. We add vocabulary as the co-design variable and sweep across
hardware ridge points.

\textbf{Getting the Most Out of Your Tokenizer (2024)}~\cite{bpemetric} provides an
analytic batch-dependent model for vocabulary cost and introduces BPB as a cross-tokenizer
quality metric. This is the closest prior art on the cost side; we extend it with
empirical measurement across hardware regimes and a full lifecycle framework.

\subsection{Roofline Model}

The roofline model~\cite{roofline} bounds the achievable performance of any operation:
\begin{equation}
\text{performance} = \min(\text{BW} \cdot I,\ \text{peak\_FLOPS})
\end{equation}
where $I$ (FLOP/byte) is arithmetic intensity. The ridge point
$\rho = \text{peak\_FLOPS} / \text{BW}$ separates memory-bound ($I < \rho$) from
compute-bound ($I > \rho$) operation. We measure $\rho$ empirically at runtime: peak
bandwidth via a large memory copy (64M fp16 elements, 128\,MB), and peak compute via
a large square GEMM ($4096 \times 4096$, fp16). This yields $\rho_\text{A10G} \approx 117$
and $\rho_\text{A100} \approx 183$ FLOP/byte, which are used to label every measurement
as memory-bound or compute-bound.

\subsection{BPB as a Quality Metric}

Bits per byte (BPB) normalizes model quality across tokenizers:
\begin{equation}
\text{BPB} = \frac{\mathcal{L}}{\ln 2} \cdot r(V)
\end{equation}
where $\mathcal{L}$ is cross-entropy loss in nats per token and $r(V)$ is tokens per
byte. Unlike raw loss, BPB is comparable across tokenizers of different vocabulary
sizes because it expresses predictive uncertainty per byte of original text,
independent of how that text was tokenized~\cite{bpemetric}.

\section{Cost Model}

\subsection{Inference Cost}

Each autoregressive decode step produces one token via:
\begin{equation}
\mathbf{logits} = \mathbf{h} W^T, \quad \mathbf{h} \in \mathbb{R}^{B \times d},\ W \in \mathbb{R}^{V \times d}
\end{equation}

The bytes moved from HBM are dominated by the weight matrix read:
$\text{bytes} \approx Vd \cdot \text{bpp}$
where bpp = 2 for fp16. The FLOPs are $2BVd$.
Arithmetic intensity (simplified for $V \gg B$):
\begin{equation}
I = \frac{2BVd}{Vd \cdot \text{bpp}} = \frac{2B}{\text{bpp}} = B \quad \text{(fp16)}
\end{equation}

Per-token head cost follows the roofline:
\begin{equation}
h(V, B) = \frac{1}{B} \cdot \max\!\left(\frac{Vd \cdot \text{bpp}}{\text{BW}},\ \frac{2BVd}{\text{peak}}\right)
\end{equation}

When $B < \rho$ (memory-bound): $h(V,B) \propto V/B$, falling linearly with batch.
When $B > \rho$ (compute-bound): $h(V,B) \propto V/\text{peak}$, independent of $B$.

The transformer body (attention + FFN, no head) contributes a V-independent cost
$c_\text{body}(B)$ measured via CUDA graph capture. Total inference cost per character:
\begin{equation}
C_\text{infer}(V, B) = \frac{c_\text{body}(B) + h(V,B)}{p(V)}
\label{eq:cinfer}
\end{equation}
where $p(V) = 1/r(V)$ is chars per token (compression, from Phase A).

This expression has a U-shape in $V$: small $V$ forces many decode steps
(small $p(V)$); large $V$ incurs large $h(V,B)$ per step. The minimum -- the
inference-optimal vocabulary -- shifts rightward as $B$ grows and $h(V,B)$
shrinks.

\subsection{Training Cost}

Training processes all tokens in the corpus once. The per-byte training cost (ms/byte):
\begin{equation}
C_\text{train}(V) = \frac{r(V)}{\tau(V)} \times 1000
\label{eq:ctrain}
\end{equation}
where $r(V)$ is tokens per byte and $\tau(V)$ is training throughput (tok/s).
Both decrease with $V$: larger $V$ compresses better (fewer tokens per byte) but
also slows each step (larger embedding and head layers). Their ratio has a minimum
at $V \approx 16$k in our experiments (Section~\ref{sec:training}).

\subsection{Lifecycle Cost and Optimal Vocabulary}

The total deployment cost over the model's lifetime:
\begin{equation}
C_\text{lifecycle}(V, B, \lambda) = C_\text{train}(V) + \lambda \cdot C_\text{infer}(V, B)
\label{eq:lifecycle}
\end{equation}

$\lambda = \text{inference\_bytes} / \text{training\_bytes}$ is a dimensionless ratio
capturing inference volume: $\lambda=1$ denotes one inference pass per training byte;
$\lambda=1000$ is typical of a production API; $\lambda=10^6$ represents a
high-traffic service (e.g., a public chatbot). Both $C_\text{train}$ and $C_\text{infer}$
are in ms/byte, making $\lambda$ a direct weight between the two costs.

The lifecycle-optimal vocabulary:
\begin{equation}
V^*(B, \lambda) = \mathop{\arg\min}_{V}\ C_\text{lifecycle}(V, B, \lambda)
\label{eq:vstar}
\end{equation}

Since both cost curves are convex (verified empirically), $V^*(B, \lambda)$
interpolates between $V^*_\text{train}$ (at $\lambda=0$) and $V^*_\text{infer}(B)$
(at $\lambda \to \infty$). The lifecycle optimum never falls outside
$[V^*_\text{train},\, V^*_\text{infer}(B)]$ -- but this interval is itself
$B$-dependent, widening dramatically at datacenter batch.

\section{Experimental Setup}
\label{sec:exp}

\subsection{Hardware and Calibration}

\begin{table}[h]
\centering
\caption{GPU hardware used in experiments. Ridge point measured empirically.}
\label{tab:hardware}
\begin{tabular}{lrrr}
\toprule
GPU & BW & Compute & Ridge \\
    & (GB/s) & (TFLOP/s) & (FLOP/byte) \\
\midrule
A10G & 473 & 60 & $\approx$117 \\
A100-SXM4 & 1359 & 252 & $\approx$183 \\
\bottomrule
\end{tabular}
\end{table}

Peak bandwidth and compute are measured at startup via \texttt{calibrate()}: a
64M-element fp16 memory copy (128\,MB, too large for L2 cache) for bandwidth, and
a $4096 \times 4096$ fp16 GEMM (intensity $\approx 1365 \gg \rho$) for peak compute.
Achieved values are 67--79\% of theoretical peak due to memory refresh cycles,
cache pressure, and thermal state. Using measured rather than spec-sheet values
yields ridge-point predictions that match observed crossover batches to within 1.6$\times$.

\subsection{Phase A: Tokenizer Training and Compression Curve}

We trained BPE tokenizers using SentencePiece~\cite{sentencepiece} on 500\,MB of
FineWeb-Edu~\cite{fineweb}, a curated English educational web corpus, at eight vocabulary
sizes: $V \in \{8192, 16384, 32768, 65536, 131072, 262144, 524288, 1048576\}$.
Training used the default BPE algorithm: starting from individual characters, iteratively
merging the most frequent adjacent token pair until reaching the target $V$.

Compression $p(V)$ (chars per token) was evaluated on a held-out 10k-line validation
split. Key result: $p(V)$ grows from 4.00 at $V=8$k to 5.19 at $V=262$k, then
saturates -- only 0.6\% additional gain from 262k to 524k, and 0.3\% from 524k to 1M.
This saturation is critical: it creates the right arm of the U-shaped $C_\text{infer}$
curve at high batch, bounding the inference optimum at $\approx$524k rather than
infinity.

\subsection{Phase B: Inference Benchmarks}

\textbf{Head measurement.} We measure the unembedding matmul
$\mathbf{h} W^T$ in isolation using CUDA events (50 timed iterations, 15 warmup,
median of 3 independent runs). Sweeping $V \in \{8\text{k}$--$1\text{M}\}$ and
$B \in \{1, 16, 64, 256, 512, 1024, 4096, 16384\}$ on a single GPU.
Measuring in isolation is essential: a full decode step launches $\sim$100 GPU kernels
(one per operation in 12 transformer layers), each incurring $\sim$15\,$\mu$s of
kernel-launch overhead. At $B=1$, this overhead ($\sim$1.5\,ms total) dwarfs the head
cost ($\sim$0.015--0.76\,ms depending on $V$), making the vocabulary signal unmeasurable.
Isolating the single head kernel removes all launch overhead.

For each (V, B) pair, we record: total latency (ms), per-token latency ($\mu$s/tok),
achieved bandwidth (GB/s), achieved compute (TFLOP/s), arithmetic intensity (FLOP/byte),
regime label (mem/compute), bandwidth utilization, and FLOP utilization.

\textbf{Body measurement.} The transformer body (12 layers, $d=512$, 8 heads, no LM head)
is timed via three methods: (1) \emph{eager mode} (baseline with full kernel-launch
overhead), (2) \emph{torch.compile} (fused kernels, reduced overhead), and
(3) \emph{CUDA graph capture} (gold standard -- the decode step is recorded once and
replayed with zero launch overhead). The gold-standard body cost at $B=1$, ctx=512 is
$c_\text{body} = 0.667$\,ms/tok. An analytical roofline estimate gives 0.065\,ms/tok,
implying the actual kernels operate at $\approx$10\% of theoretical efficiency -- typical
for small-batch transformer inference without custom kernels.

\subsection{Phase C: Training and Quality Measurement}
\label{sec:training}

\textbf{100M-scale run.}
We trained 6 GPT models with identical architecture ($d=512$, 12 layers, 8 heads,
ctx=2048) at $V \in \{8192, 16384, 32768, 65536, 131072, 262144\}$ for 10k steps
each, batch=16, AdamW (lr=3e-4, cosine schedule), on 50M tokens of FineWeb-Edu.
All 6 models share the same transformer body (37.7M parameters); only the embedding
and head layers grow with $V$. Hardware: 8$\times$ A100 with DataParallel.

A key implementation challenge: the standard DataParallel approach gathers the full
$[\text{batch} \times \text{ctx} \times V]$ logit tensor on GPU 0 before computing
the cross-entropy loss. At $V=65536$, this tensor alone requires
$16 \times 2048 \times 65536 \times 4 = 8$\,GB, causing out-of-memory errors.
We resolve this by wrapping the model in a \texttt{GPTWithLoss} module that computes
the loss inside each GPU shard and gathers only the scalar loss to GPU 0, reducing
cross-GPU communication from gigabytes to bytes.

\textbf{1.3--2.3B scale run (scale test).}
To test whether quality differentiates across $V$ at larger scale, we trained 3 models
at $V \in \{16384, 65536, 262144\}$ with architecture $d=2048$, 24 layers, 16 heads,
ctx=1024, for 5k steps, batch=32 total across 8$\times$ A100.

At this scale, DataParallel is insufficient: the 2.3B model at $V=262$k requires
$\approx$19\,GB of AdamW optimizer states (momentum + variance + master weights),
all of which DataParallel places on GPU 0, saturating its 40\,GB HBM.
We switch to \textbf{Fully Sharded Data Parallel (FSDP)}, which shards parameters,
gradients, and optimizer states evenly across all 8 GPUs, reducing per-GPU optimizer
memory from $\sim$19\,GB to $\sim$2.4\,GB.

Training was initially attempted with fp16 mixed precision. At $V=262144$, the raw
unembedding logits can exceed fp16's maximum representable value ($\approx$65504),
causing overflow and training divergence (loss increasing monotonically from $\sim$9
to $>$18 over 2k steps). Switching to \textbf{bf16} (same exponent range as fp32,
maximum $\approx 3.4 \times 10^{38}$) eliminates the overflow. A learning rate of
6e-5 (vs.\ 3e-4 at 100M scale) is used to stabilize 2B+ parameter training.

\section{Results}

\subsection{Regime Crossover Confirms Roofline Prediction}

Table~\ref{tab:crossover} shows the batch at which the unembedding transitions from
memory-bound to compute-bound on each GPU. The measured crossover matches the
theoretical ridge point to within a factor of 1.5--2$\times$ (the small discrepancy
arises because the full bytes-moved includes activations and logits, not just the weight
matrix).

\begin{table}[h]
\centering
\caption{Measured memory-to-compute crossover batch vs.\ predicted from ridge point.}
\label{tab:crossover}
\begin{tabular}{lrrr}
\toprule
GPU & Ridge & Predicted $B$ & Measured $B$ \\
\midrule
A10G & 117 & $\approx$117 & 128--256 \\
A100 & 183 & $\approx$183 & 256--512 \\
\bottomrule
\end{tabular}
\end{table}

\begin{figure}[h]
\centering
\includegraphics[width=\columnwidth]{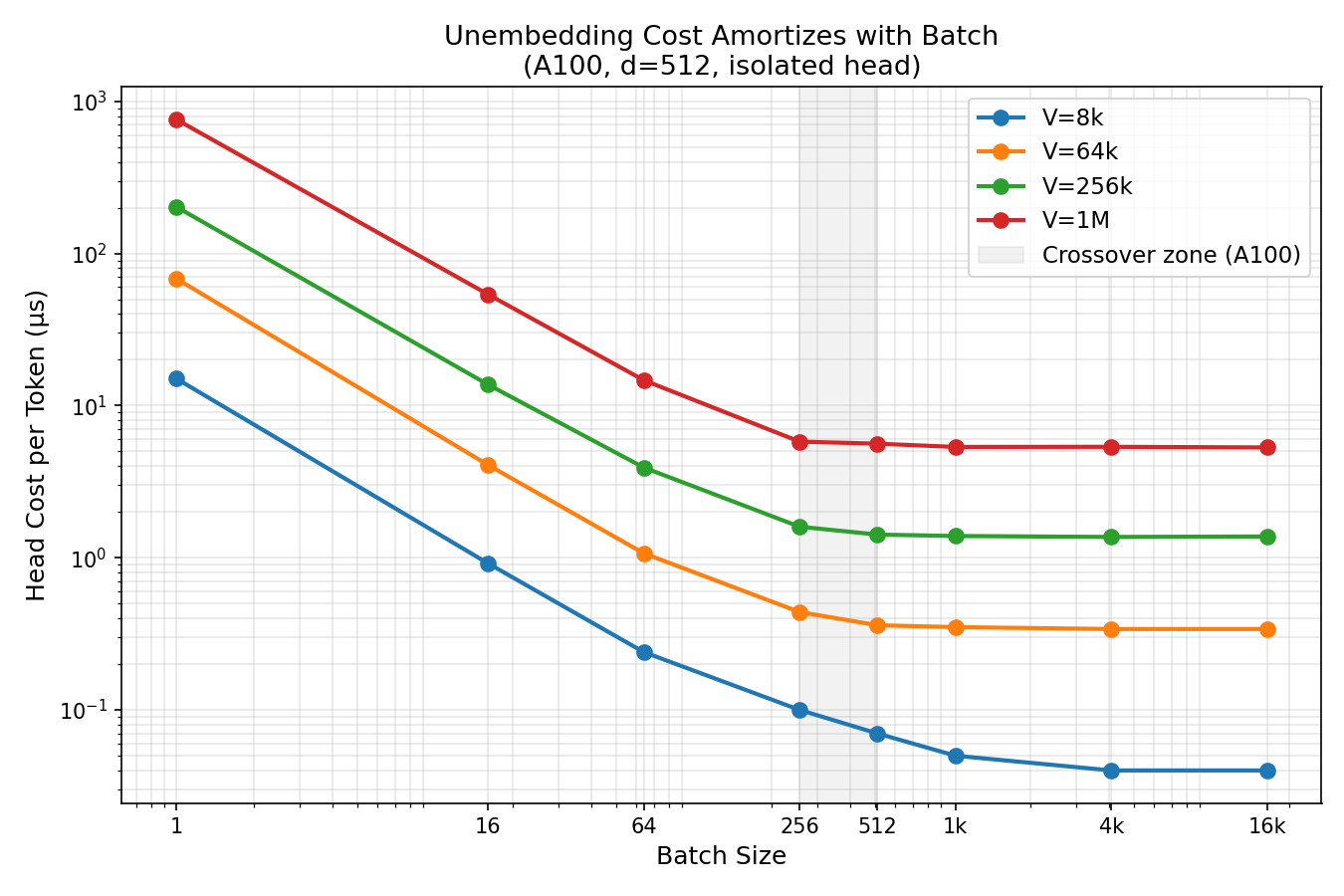}
\caption{Per-token unembedding cost vs.\ batch on A100 for four vocabulary sizes.
Gray band marks the memory-to-compute crossover. From $B=1$ to $B=16384$, cost
drops 143--377$\times$ depending on $V$.}
\label{fig:amortization}
\end{figure}

Fig.~\ref{fig:amortization} shows per-token head cost vs.\ batch on A100.
At $B=1$, bandwidth utilization is 82--99\% and FLOP utilization is $<$1\%
(memory-bound). By $B=512$, FLOP utilization reaches 49--97\% (compute-bound).
The crossover shifted from $B \approx 128$--256 on A10G to $B \approx 256$--512 on A100,
tracking the ridge ratio $183/117 \approx 1.6\times$. This confirms the effect is
roofline physics, not an artifact of one platform.

\subsection{Inference-Optimal Vocabulary Shifts 16$\times$ with Batch}

Table~\ref{tab:head_data} shows per-token head cost at key (V, B) pairs on A100.
The amortization ratio (column 5) ranges from 143$\times$ to 377$\times$ from
$B=1$ to $B=16384$, confirming the theoretical $\propto B$ scaling.

\begin{table}[h]
\centering
\caption{Per-token head cost ($\mu$s) on A100 and amortization ratio $B=1 \to B=16384$.}
\label{tab:head_data}
\begin{tabular}{lrrrr}
\toprule
$V$ & $B=1$ & $B=64$ & $B=4096$ & Ratio \\
\midrule
8192    &  15.1 & 0.24 & 0.04 & 377$\times$ \\
65536   &  68.4 & 1.07 & 0.34 & 201$\times$ \\
262144  & 202.7 & 3.92 & 1.37 & 148$\times$ \\
1048576 & 760.8 & 14.70 & 5.31 & 143$\times$ \\
\bottomrule
\end{tabular}
\end{table}

Using $c_\text{body}=0.667$\,ms/tok (CUDA graph, ctx=512, $B=1$) and real BPE
compression from Phase A, the cost-optimal vocabulary from Eq.~(\ref{eq:cinfer}):

\begin{itemize}
\item $B=1$: $V^*_\text{infer} = 32768$. Head cost at $V=262$k is 203\,$\mu$s/tok vs.\
  body 667\,$\mu$s/tok. Head contributes 23\%, creating a real penalty for large $V$.
\item $B=64$: $V^*_\text{infer} = 524288$. Head at $V=262$k is now 61\,$\mu$s/tok
  vs.\ body 75\,$\mu$s/tok. Head is 45\% of total but compression gain past 262k is
  only 0.6\%, so the minimum is at 524k where compression saturates.
\end{itemize}

\begin{figure}[h]
\centering
\includegraphics[width=\columnwidth]{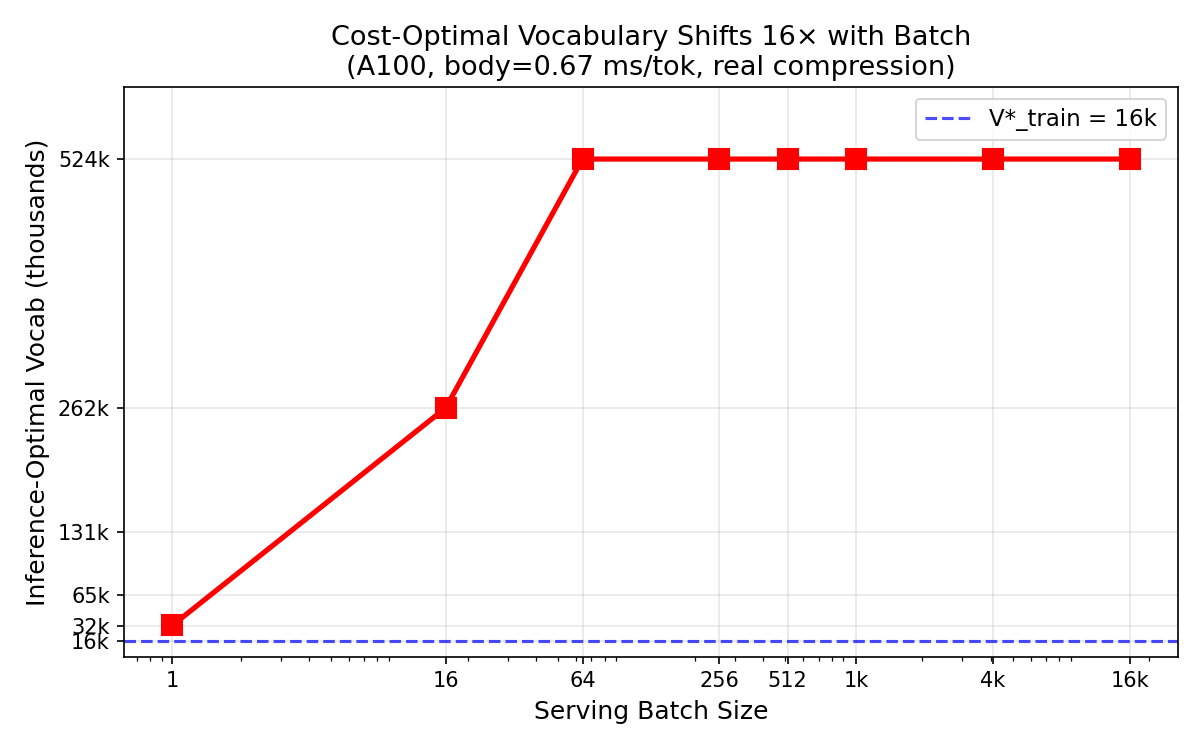}
\caption{Inference-cost-optimal vocabulary vs.\ serving batch (A100, real BPE compression,
$c_\text{body}=0.667$\,ms/tok). The blue dashed line shows $V^*_\text{train}=16$k.}
\label{fig:optimal_v}
\end{figure}

The 16$\times$ shift ($32$k $\to$ $524$k) is shown in Fig.~\ref{fig:optimal_v}.
The saturation of the BPE compression curve at $\approx$524k (Phase A) naturally
bounds the optimum, preventing it from running to infinity at high batch.

\subsection{Training Results and Training-Optimal Vocabulary}

Table~\ref{tab:training} summarizes Phase C results at 100M scale.

\begin{table}[h]
\centering
\caption{Training results at 100M scale (d=512, 12L, 10k steps, 8x A100).}
\label{tab:training}
\begin{tabular}{lrrrr}
\toprule
$V$ & BPB & tok/s & $C_\text{train}$ (ms/byte) & Params \\
\midrule
8192    & 1.368 & 198,401 & 1.26e-3 & 47.3M \\
16384   & 1.344 & 191,215 & \textbf{1.18e-3} & 55.7M \\
32768   & 1.347 & 174,990 & 1.20e-3 & 72.4M \\
65536   & 1.346 & 151,934 & 1.32e-3 & 106.0M \\
131072  & 1.355 & 119,272 & 1.64e-3 & 173.1M \\
262144  & 1.355 &  83,132 & 2.32e-3 & 307.3M \\
\bottomrule
\end{tabular}
\end{table}

Two observations: (1) BPB is flat across the full $V$ range ($<$2\% spread, best at
$V=16$k), confirming that vocabulary is a pure systems decision at this model scale
with negligible quality impact. (2) Training throughput drops 2.4$\times$ from $V=8$k
to $V=262$k, reflecting the larger embedding/head layers. The training-cost minimum is
$V^*_\text{train} = 16384$: below this, poor compression forces more tokens per byte;
above this, the throughput penalty outweighs compression gains.

\subsection{Quality Shifts Rightward at Scale}

At 1.3--2.3B scale (FSDP, bf16, 5k steps), BPB results are:

\begin{table}[h]
\centering
\caption{Quality at 1.3--2.3B scale (d=2048, 24L, FSDP, bf16, 5k steps).}
\label{tab:largescale}
\begin{tabular}{lrrr}
\toprule
$V$ & BPB & Params & tok/s \\
\midrule
16384  & 1.399 & 1.28B & 120,343 \\
65536  & \textbf{1.387} & 1.48B & 112,415 \\
262144 & 1.397 & 2.28B &  88,865 \\
\bottomrule
\end{tabular}
\end{table}

\begin{figure}[h]
\centering
\includegraphics[width=\columnwidth]{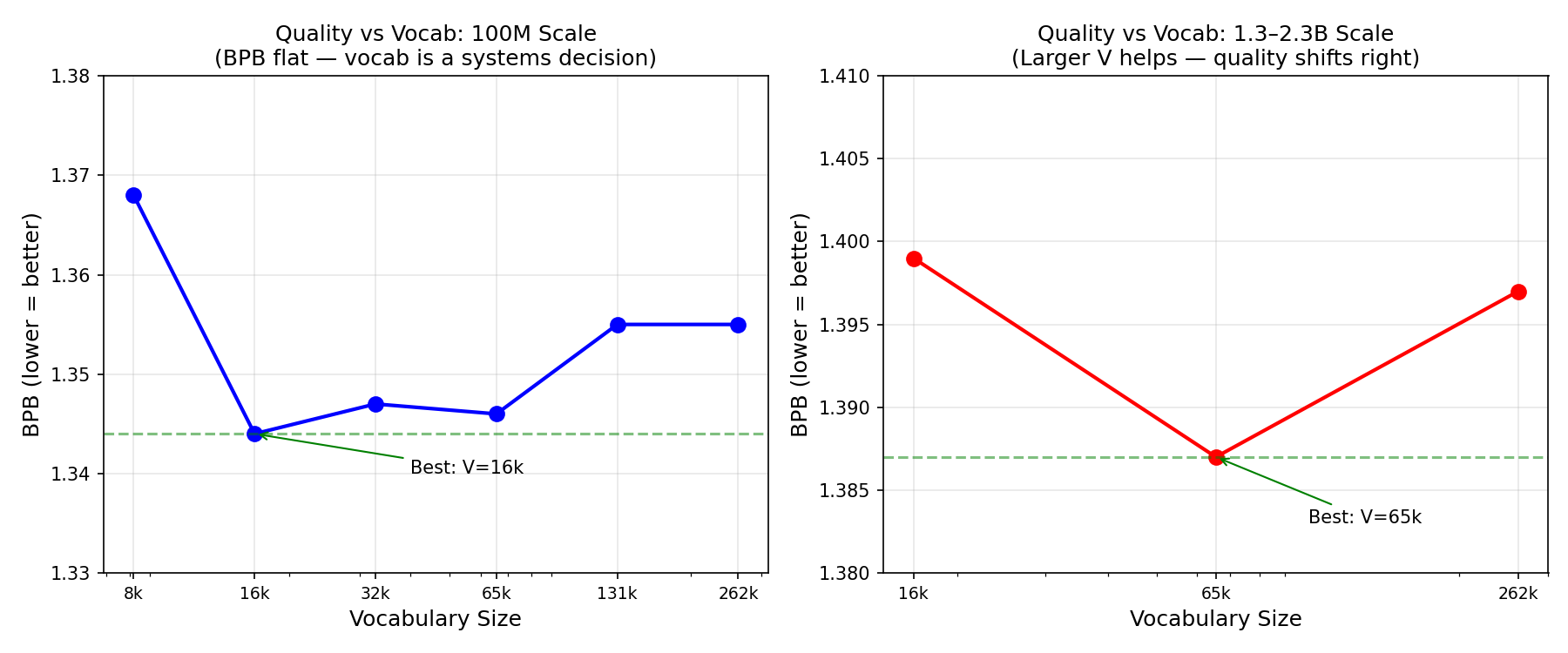}
\caption{BPB vs.\ vocabulary at two scales. Left (100M): BPB is flat; quality
does not constrain vocabulary choice. Right (1.3--2.3B): the optimum shifts to
$V=65$k, confirming scale-dependent vocabulary preference.}
\label{fig:bpb}
\end{figure}

The quality optimum shifted from $V=16$k (at 100M) to $V=65$k (at 1.5B): BPB improves
from 1.399 to 1.387 (0.9\% gain), consistent with Tao et al.'s prediction of
$V^* = k\sqrt{N} \approx 63$k at $N=1.5$B.
$V=262$k is slightly worse than $V=65$k, likely because at 2.3B total parameters the
embedding and head account for $\sim$1.1B of 2.3B parameters, leaving less capacity
for the transformer body that performs language modeling. Fig.~\ref{fig:bpb}
shows BPB vs.\ vocabulary at both scales.

\subsection{Lifecycle Sweep}

Table~\ref{tab:lifecycle} presents $V^*(B, \lambda)$ from Eq.~(\ref{eq:vstar}),
computed using measured $C_\text{train}$ (Table~\ref{tab:training}) and
$C_\text{infer}$ (Section IV-B). At $\lambda=0$ (training only), $V^*=16$k
regardless of batch. Any serving volume ($\lambda \geq 1$) immediately shifts $V^*$
away from the training optimum. The transition is gradual: at $B=256$, $\lambda=1$
gives $V^*=65$k but reaching $V^*=262$k requires $\lambda \geq 100$, corresponding
to serving at least 100 bytes of inference per byte of training data.

\begin{table}[h]
\centering
\caption{Lifecycle-optimal vocabulary $V^*(B,\lambda)$.
Values in thousands (e.g., 262 = 262k tokens).
$\lambda=1000$ is typical production.}
\label{tab:lifecycle}
\setlength{\tabcolsep}{4pt}
\begin{tabular}{lcccccc}
\toprule
$B$ & $\lambda{=}0$ & $\lambda{=}1$ & $\lambda{=}10$ & $\lambda{=}100$ & $\lambda{=}10^3$ & $\lambda{=}\infty$ \\
\midrule
1   & 16 & 32 &  32 &  32 &  32 &  32 \\
16  & 16 & 65 &  65 &  65 &  65 &  65 \\
64  & 16 & 65 & 131 & 131 & 131 & 131 \\
256 & 16 & 65 & 131 & 262 & 262 & 262 \\
1k+ & 16 & 65 & 131 & 262 & 262 & 262 \\
\bottomrule
\end{tabular}
\end{table}

\section{Infrastructure Implications}

\subsection{Capacity Planning Impact}

Vocabulary size directly affects the number of accelerators required to meet a
throughput SLA. The inference cost per character (Eq.~\ref{eq:cinfer}) determines
how many decode steps the cluster must execute per unit of output text. Moving from
the conventional $V=32$k to the lifecycle-optimal $V=262$k at datacenter batch
($B=256$, $\lambda \geq 100$) reduces per-character inference cost by $\approx$2$\times$,
directly halving the number of GPU-hours required for a fixed output volume.

Concretely, at $B=64$: $V=524$k reduces head cost from 61\,$\mu$s/tok ($V=262$k)
and provides 0.6\% more compression. The combined effect on cost-per-character is
a $\approx$10\% reduction vs.\ $V=262$k. While modest, at hyperscale inference
volumes this translates to substantial cluster cost savings.

Vocabulary also interacts with KV-cache utilization. Larger $V$ produces shorter
token sequences for the same text (better compression), which directly reduces
KV-cache size: a 16$\times$ more compressed tokenizer halves sequence length, halving
KV-cache memory requirements. For long-context serving, this can be the dominant
capacity driver.

\subsection{Theoretical Extension to Quantization}

While we did not run quantization experiments, the roofline model predicts the
effect analytically. int4 quantization of the unembedding (bpp = 0.5 instead of 2)
would reduce bytes read from HBM by 4$\times$, shifting arithmetic intensity from
$I \approx B$ to $I \approx 4B$ and raising the effective ridge point 4$\times$.
The memory-to-compute crossover batch would shift from $B \approx 256$ to
$B \approx 64$ on A100, meaning the head amortizes at lower batch sizes and the
lifecycle optimum moves toward larger $V$ even at modest batch. Empirical
validation of this prediction is left as future work.

\subsection{Provisioning Recommendations}

For infrastructure teams making vocabulary selection decisions:

\begin{itemize}
\item \textbf{On-device / edge ($B=1$):} Use $V \approx 32$k. The head is memory-bound
  and scales linearly with $V$; the optimal vocabulary minimizes memory traffic per step
  while maintaining sufficient compression. Quality is unaffected ($<$1\% BPB difference
  across 32k--262k at 1.5B scale).

\item \textbf{API servers ($B=16$--64):} Use $V \approx 65$--131k. The head begins to
  amortize; compression reduces total step count. The lifecycle optimum at $\lambda=10$
  (moderate serving) falls in this range.

\item \textbf{Datacenter / high-throughput ($B \geq 256$, $\lambda \geq 100$):} Use
  $V \approx 262$k. Full head amortization; maximum compression gain; no quality penalty
  at $\geq$1B model scale. This is 8--16$\times$ larger than the training convention
  of 16--32k.
\end{itemize}

These recommendations apply at model scales $\geq$1B where quality weakly favors
larger $V$. At $<$300M parameters, quality is flat and the choice reduces to the
training-inference cost trade-off alone.

\section{Limitations}

\textbf{Model scale.} Our largest training run is 2.3B parameters. Tao's scaling law
$V^* \propto N^{0.5}$ predicts the quality benefit of large $V$ continues to grow with
scale; at 7B+ the quality optimum would be $\approx$150k+ (untested here).

\textbf{Training budget.} We trained 1.3--2.3B models for only 5k steps. The $V=262$k
model may be undertrained relative to $V=65$k at an equal compute budget (FLOPs), since
the larger embedding/head consume parameters that would otherwise contribute to the
transformer body.

\textbf{Corpus.} Tokenizers were trained on English FineWeb-Edu. Compression curves
differ for code (higher compression from repeated patterns), multilingual text (lower
per-character compression for non-Latin scripts), and mathematical notation.
The optimal $V$ values are corpus-dependent.

\textbf{Architecture.} All results use decoder-only GPT. Mixture-of-experts (MoE) models,
state-space models, and encoder-decoder architectures have different body/head cost
ratios and may exhibit different optimal vocabulary sizes.

\section{Conclusion}

We have shown that the cost-optimal tokenizer vocabulary is not a constant but a
deployment-regime-dependent infrastructure parameter. The lifecycle framework
$C_\text{lifecycle}(V, B, \lambda)$ quantifies a three-way tension: training cost
is minimized at $V \approx 16$k; quality (BPB) is minimized at $V \approx 65$k
at 1.5B scale; and inference cost at serving batch is minimized at $V \approx 262$k+
for $B \geq 64$.

The key practical finding: for any production LLM deployment with $\lambda \geq 100$
and $B \geq 64$, the lifecycle-optimal vocabulary is 8--16$\times$ larger than
training convention suggests, with $<$2\% quality cost in the measured parameter range.
The effect is roofline physics -- confirmed on two GPU ridge points and predictable
from hardware specs -- making it applicable to future GPU generations without
re-measurement.

For infrastructure teams, this result is directly actionable: vocabulary selection
should be treated as a capacity planning input alongside batch size, serving regime,
and quantization strategy, rather than a fixed convention inherited from training.

\section*{Acknowledgment}

The authors thank the open-source communities behind PyTorch, SentencePiece, and
HuggingFace Datasets. Experiments were conducted on AWS EC2 p4d.24xlarge (8$\times$
A100-SXM4-40GB) and g5.4xlarge (1$\times$ A10G-24GB) instances.

\textbf{AI Tool Disclosure (IEEE Policy):} This work made substantial use of Claude
(Anthropic) as an AI assistant throughout the research process. The authors originated
the core research idea -- that vocabulary size should be treated as a deployment-regime
parameter rather than a training-time constant -- and defined the experimental questions
and hypotheses. Claude was used to: (1) conduct the literature survey, identifying and
summarizing relevant prior work (Tao 2024, Length-MAX, Compute-Optimal Tokenization,
Hardware Co-Design Scaling Laws, and others) and assessing novelty and differentiation
of our contribution; (2) implement the experimental harness
(\texttt{bench\_head.py}, \texttt{bench\_body.py}, \texttt{train\_models.py},
\texttt{lifecycle\_sweep.py} and related scripts) from author-provided specifications
and seed ideas; (3) debug training instabilities (fp16 overflow, DataParallel OOM,
FSDP configuration); (4) assist with data analysis and result interpretation;
and (5) draft and revise sections of this manuscript. All experimental design decisions,
scientific claims, and final interpretations are the authors' own. The authors verified
all code, reviewed all outputs, and take full responsibility for the content of this
paper.


\end{document}